\documentclass{article}
\usepackage{spconf,amsmath,graphicx}
\usepackage{booktabs} 
\usepackage{url}
\usepackage{enumitem}
\setlist{nosep, leftmargin=14pt}
\usepackage{mwe} 

\title{Metadata-Aligned 3D MRI Representations for Contrast Understanding and Quality Control}

\name{
\begin{tabular}{c}
Mehmet Yigit Avci$^{\star}$, Pedro Borges$^{\star}$, Virginia Fernandez$^{\star}$, Paul Wright$^{\star}$,\\
Mehmet Yigitsoy$^{\dagger}$, Sebastien Ourselin$^{\star}$, Jorge Cardoso$^{\star}$
\end{tabular}
}

\address{
$^{\star}$School of Biomedical Engineering \& Imaging Sciences, King’s College London, London, UK \\
$^{\dagger}$deepc GmbH, Munich, Germany
}
\begin{document}
%
\maketitle
\begin{abstract}
Magnetic Resonance Imaging suffers from substantial data heterogeneity and the absence of standardized contrast labels across scanners, protocols, and institutions, which severely limits large-scale automated analysis. A unified representation of MRI contrast would enable a wide range of downstream utilities, from automatic sequence recognition to harmonization and quality control, without relying on manual annotations. To this end, we introduce MR-CLIP, a metadata-guided framework that learns MRI contrast representations by aligning volumetric images with their DICOM acquisition parameters. The resulting embeddings can unsupervisedly cluster MRI sequences and outperform supervised 3D baselines under data scarcity in few-shot sequence classification. Moreover, MR-CLIP enables unsupervised data quality control by identifying corrupted or inconsistent metadata through image–metadata embedding distances. By transforming routinely available acquisition metadata into a supervisory signal, MR-CLIP provides a scalable foundation for label-efficient MRI analysis across diverse clinical datasets.
\end{abstract}
\section{Introduction}
\label{sec:intro}

Magnetic Resonance Imaging (MRI) is indispensable in modern clinical practice, providing unparalleled soft-tissue contrast and diagnostic flexibility through diverse acquisition protocols and pulse sequences. However, this versatility introduces significant challenges for automated analysis, as clinical datasets exhibit substantial heterogeneity arising from differences in scanner manufacturers, field strengths, and patient-specific acquisition settings. Such variability complicates data organization and undermines the reliability of automated processing pipelines \cite{sinha2024mrqa,gauriau2020dicom}. To mitigate these challenges, previous studies have utilized DICOM \cite{DICOMStandard} acquisition metadata for tasks such as sequence detection \cite{liang2021mri,gauriau2020dicom} and metadata-based quality control \cite{mrqy,qc}. Beyond these applications, metadata has also proven valuable for harmonization across scanners and protocols \cite{tumsyn}, and more generally as a guiding signal for robust image analysis \cite{contextmri,oct-clip}. Building on these metadata-driven approaches, we extend the 2D MR-CLIP framework \cite{avci2025mrclipefficientmetadataguidedlearning}, which aligns individual slices with their metadata, into a fully 3D model that captures volumetric context across entire scans. Inspired by \cite{tumsyn, radford2021learning}, MR-CLIP converts structured DICOM metadata into natural language templates and learns to contrastively align them with corresponding MRI volumes. This unsupervised training produces rich and contrast-aware embeddings that capture underlying physics of each acquisition. Importantly, the framework provides a single representation that supports a wide range of downstream tasks: retrieval of images or metadata (critical for organizing large datasets), sequence classification and automatic data quality control (QC). Our main contributions are three-fold:
\begin{itemize}
    \item We propose a 3D metadata-guided contrastive learning framework that disentangles image contrast from anatomical variability, producing contrast representations across full MRI volumes.
    \item The learned embeddings enable accurate few-shot sequence classification, outperforming 3D CNNs in low-data settings, and naturally cluster by sequence, highlighting their quality and encoding fidelity.
    \item We introduce a novel multimodal embedding–based method for unsupervised MRI QC, where dissimilarity between image and metadata embeddings indicates missing or corrupted DICOM tags, enabling scalable evaluation of large imaging datasets.
\end{itemize}

\begin{figure*}[t]
    \centering
    \includegraphics[width=\textwidth]{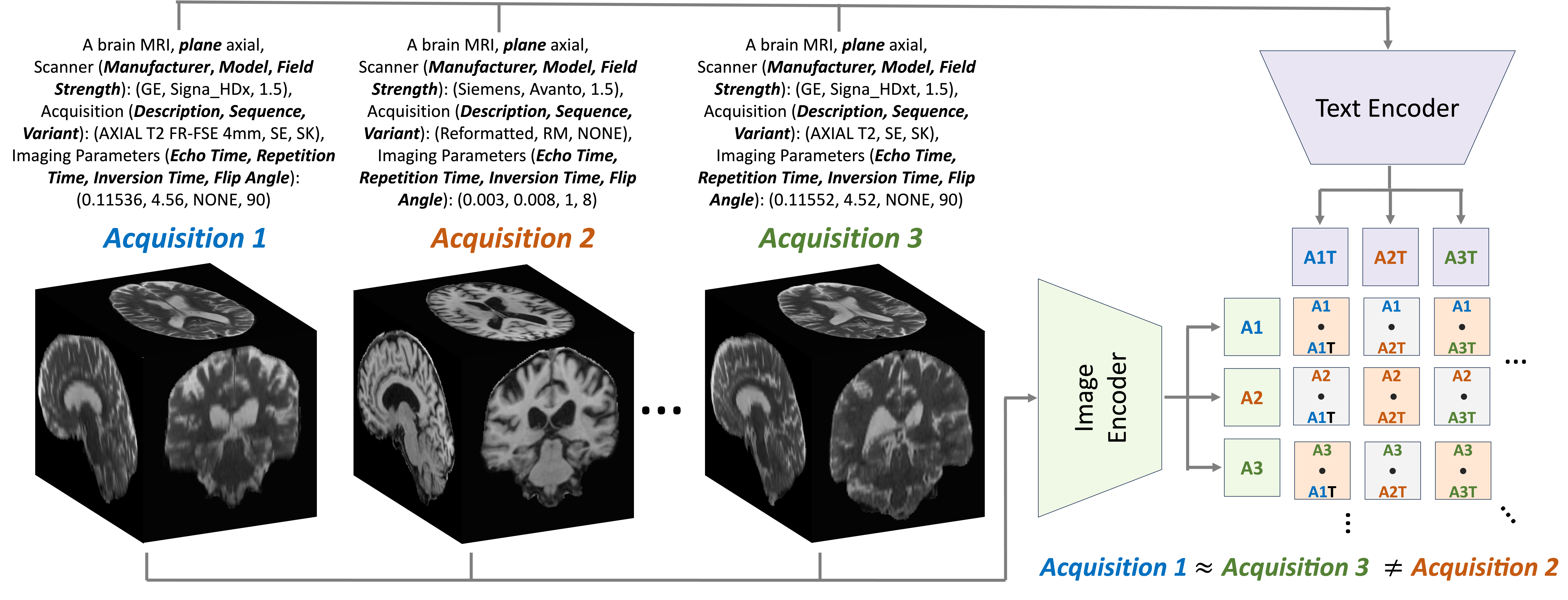}
    \caption{MR-CLIP aligns MRI volumes with their corresponding DICOM metadata through contrastive learning. A 3D image encoder and a metadata encoder jointly learn to associate similar acquisitions while distinguishing different contrasts, resulting in contrast-aware representations that are robust to anatomical and subtle acquisition variability.}
    \label{fig:fig1}
    
\end{figure*}

\section{Methods}
\label{sec:method}

MR-CLIP learns metadata-aligned MRI representations by contrastively matching volumetric image embeddings with structured DICOM metadata, as illustrated in Fig.~\ref{fig:fig1}. For each acquisition, volumetric features are extracted using a 3D image encoder, and the associated acquisition parameters are converted into natural language templates and projected by a text encoder into a shared embedding space. To minimize the impact of minor acquisition variations that do not meaningfully alter image contrast, we follow the same approach in \cite{avci2025mrclipefficientmetadataguidedlearning} and group scans with similar imaging parameters. Specifically, numeric fields (\textit{Echo Time}, \textit{Repetition Time}, \textit{Inversion Time}) are jointly quantized into a 20×20 grid, while categorical fields (\textit{Manufacturer}, \textit{Scanner Model}, \textit{Imaging Plane}, \textit{Field Strength}, \textit{Sequence Type}, \textit{Sequence Variant}, \textit{Series Description}, \textit{Flip Angle}) are grouped by unique combinations. This process yields semantically consistent acquisition clusters that reflect true contrast-level distinctions rather than trivial parameter differences. MR-CLIP is trained using a Supervised Contrastive (SupCon) loss \cite{supcon}. Let $z_i$ denote the anchor embedding for sample $i$, and let $P(i)$ be the set of positive embeddings for $i$, including exact matches and other samples from the same metadata group. The loss for anchor $i$ is
\begin{equation}
\mathcal{L}_i = -\frac{1}{|P(i)|} \sum_{p \in P(i)} 
\log \frac{\exp(z_i^\top z_p / \tau)}{\sum_{a \in A(i)} \exp(z_i^\top z_a / \tau)}
\end{equation}
where $A(i)$ is the set of all embeddings in the batch excluding $i$, $z$ represents any image or metadata embedding, and $\tau$ is a temperature hyperparameter. Final loss is given as follows:
\begin{equation}
\mathcal{L} = \frac{1}{2} \left( \mathcal{L}_{\text{SupCon}}^{\text{img} \rightarrow \text{text}} + \mathcal{L}_{\text{SupCon}}^{\text{text} \rightarrow \text{img}} \right)
\end{equation} 
Compared to standard InfoNCE \cite{infonce}, which considers only a single positive per anchor, SupCon naturally handles multiple positives, encouraging the model to cluster semantically similar acquisitions. 
\subsection{Data and Implementation Details}
We use a large-scale dataset of 3D brain MRIs from King’s College Hospital (KCH) and Guy’s and St Thomas’ NHS Foundation Trust (GSTT), comprising 40,005 subjects and 169,634 volumes. These scans include 21,660 unique acquisition configurations derived from DICOM metadata, which are grouped into 1,415 contrast categories with our grouping strategy. The dataset is divided into training sets (60\%), validation sets (10\%), and test sets (30\%) at the scan level. All scans are rigidly registered to MNI space and skull-stripped with SynthStrip \cite{synthstrip}.

MR-CLIP is implemented in PyTorch and trained on three NVIDIA A100 GPUs (40 GB) using a per-GPU batch size of 150 with sharded loss following \cite{open_clip}. The model is optimized with Adam ($\text{lr}=1\mathrm{e}{-4}$) and a weight decay of 0.2 for 100 epochs, including 2,000 warm-up steps. Gradient checkpointing is used to reduce memory consumption. 

\section{Results}
\label{sec:results}
We structure the validation of our volumetric MR-CLIP framework into three complementary stages designed to assess its representational quality, and clinical applicability. First, we evaluate representation quality through linear contrast classification to measure how effectively the model encodes semantic imaging properties across 2D and 3D architectures. Second, we assess sequence recognition capabilities by analyzing the learned embedding space through t-SNE and few-shot classification. Finally, we demonstrate the clinical utility of the framework by applying it for unsupervised QC, where the model identifies simulated DICOM field corruptions using cross-modal embedding distances.

\begin{figure}[htb]

\begin{minipage}[b]{1.0\linewidth}
  \centering
  \centerline{\includegraphics[width=8.5cm]{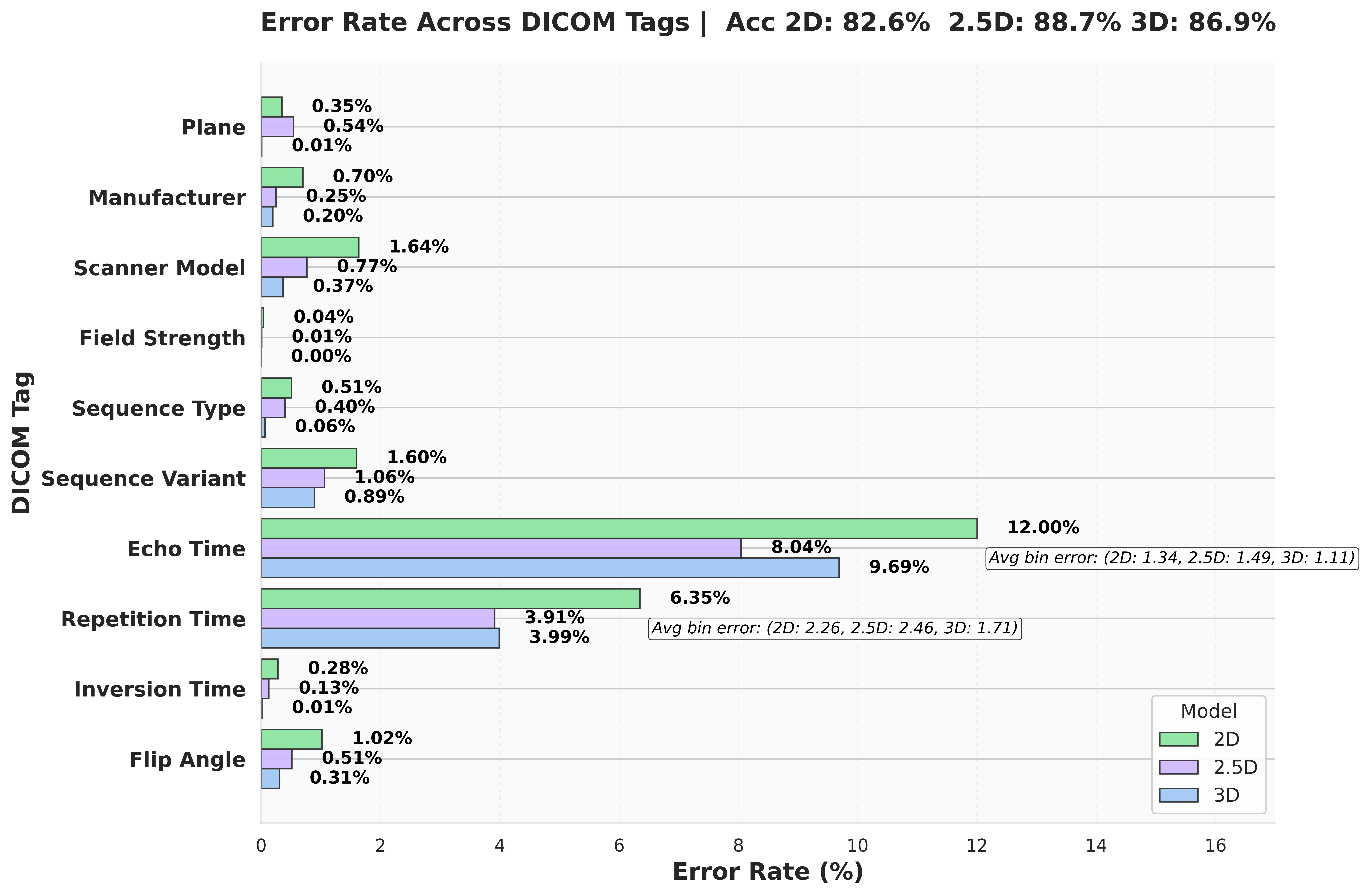}}
  \medskip
\end{minipage}
\caption{Error rates across DICOM tags based on linear probe classification results.}
\label{fig:error_rate}
\end{figure}

As shown in Fig. \ref{fig:error_rate}, we evaluate linear probe classification results and individual error rates across DICOM tags using 2D, 2.5D (aggregated slice-level results), and 3D MR-CLIP variants. The 2.5D model achieves the highest overall accuracy (88.7\%), with 3D achieving comparable performance (86.9\%), suggesting that aggregating local spatial context across slices provides an effective balance between representational richness and efficiency. Discrete tags such as \textit{Acquisition Plane} and \textit{Field Strength} are predicted with near-perfect accuracy, demonstrating robust encoding of categorical metadata. In contrast, continuous parameters like TE and TR exhibit higher misclassification rates due to discretization, though average bin-level deviations remain small, indicating that predictions remain close to the true values even when not exact.

\begin{figure}[htb]

\begin{minipage}[b]{1.0\linewidth}
  \centering
  \centerline{\includegraphics[width=8.5cm]{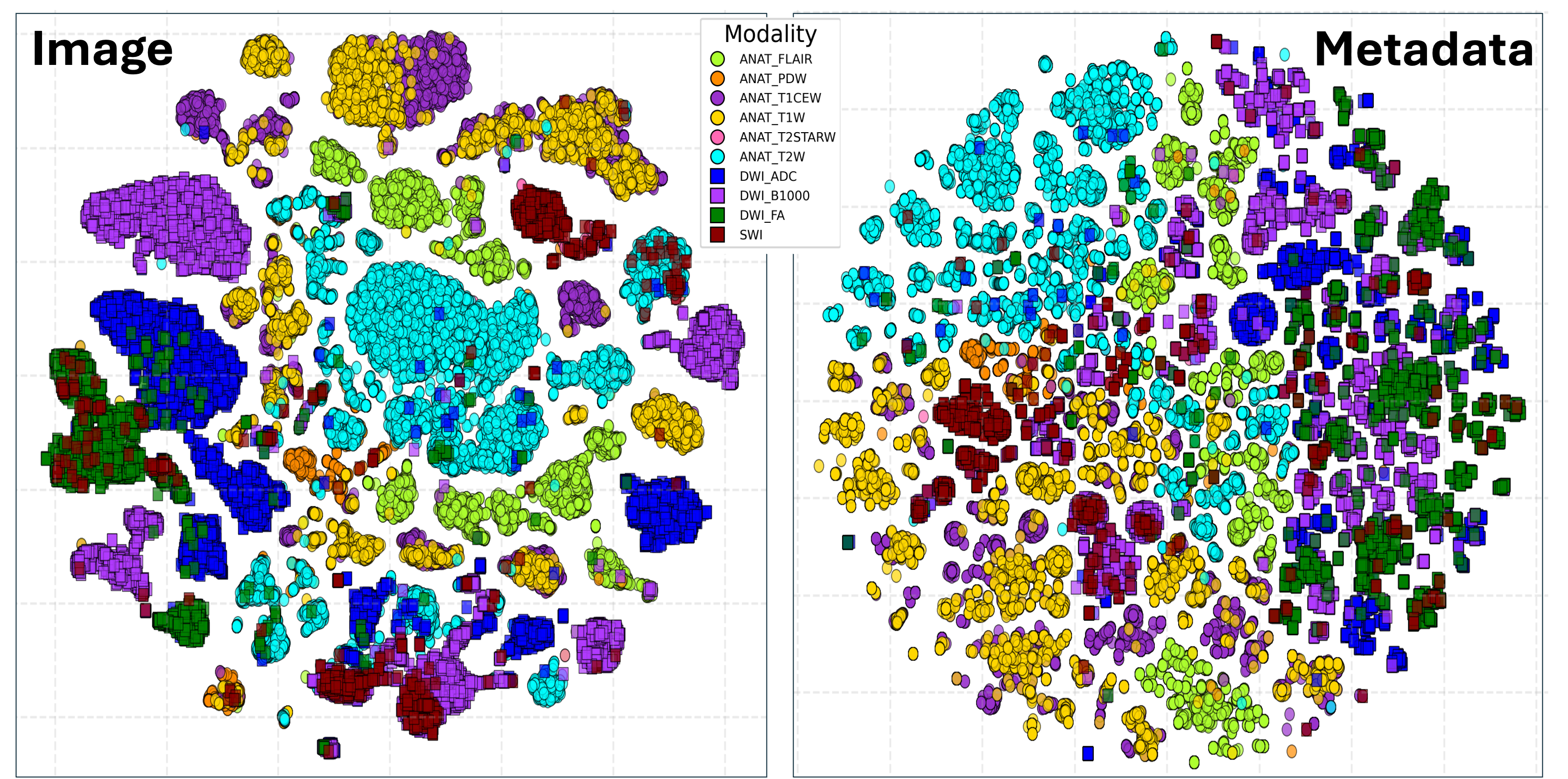}}
  \medskip
\end{minipage}
\caption{t-SNE visualizations of image and metadata embeddings, color coded by sequence.}
\label{fig:tsne}
\end{figure}

\begin{figure}[htb]

\begin{minipage}[b]{1.0\linewidth}
  \centering
  \centerline{\includegraphics[width=8.5cm]{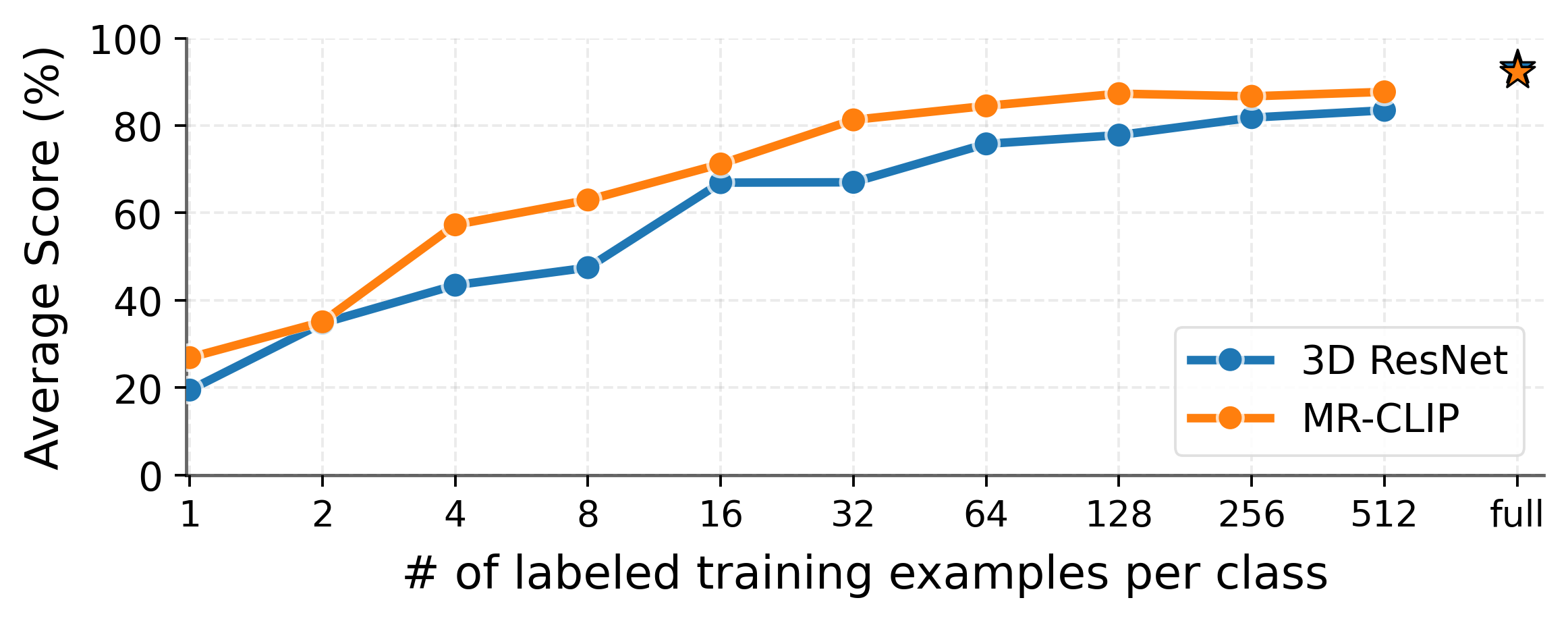}}
  \medskip
\end{minipage}
\caption{Few-shot learning performance of linear classifier trained on image embeddings of MR-CLIP, compared to supervised 3D ResNet baseline.}
\label{fig:few_shot}
\end{figure}

The t-SNE visualizations (Fig. \ref{fig:tsne}) reveal distinct clusters for different MRI sequence types, demonstrating semantically meaningful contrast embeddings that are independent from anatomical variation. This structured latent space supports efficient generalization, as confirmed by few-shot sequence classification (Fig. \ref{fig:few_shot}). Across low-shot regimes, linear classifier trained on image embeddings of MR-CLIP consistently outperforms the supervised 3D ResNet, while performance converges in the fully supervised setting. These results highlight that unsupervised metadata-guided pre-training provides an effective initialization, particularly valuable in clinical scenarios with limited labeled data.

\begin{table}[t]
\centering
\caption{Synthetic Metadata Corruptions for Unsupervised Quality Control}
\label{tab:qc_methodology}
\setlength{\tabcolsep}{3pt} 
\renewcommand{\arraystretch}{0.9} 
\begin{tabular}{p{0.18\columnwidth} p{0.12\columnwidth} p{0.58\columnwidth}}
\toprule
\textbf{Type} & \textbf{Level} & \textbf{Description} \\
\midrule
\textbf{Numeric} & Small & Slight scaling within normal range. \\
 \textbf{Error (TE, TR,}& Med. & Shift beyond expected sequence range (mimics another sequence). \\
 \textbf{TI)}& Large & Unit error (e.g., s$\rightarrow$ms, $\times$1000). \\
\midrule
\textbf{Wrong} & Small & Wrong \textit{Series Description}. \\
 \textbf{Tag}& Med. & + Wrong \textit{Sequence Info}. \\
 & Large & + Wrong \textit{Scanner Info}. \\
\midrule
\textbf{Missing} & Small & Missing \textit{Series Description}. \\
 \textbf{Tag}& Med. & + Missing \textit{Sequence Info}. \\
 & Large & + Missing \textit{Scanner Info}. \\
\bottomrule
\end{tabular}
\end{table}

\begin{figure}[htb]
\begin{minipage}[b]{1.0\linewidth}
  \centering
  \centerline{\includegraphics[width=8.5cm]{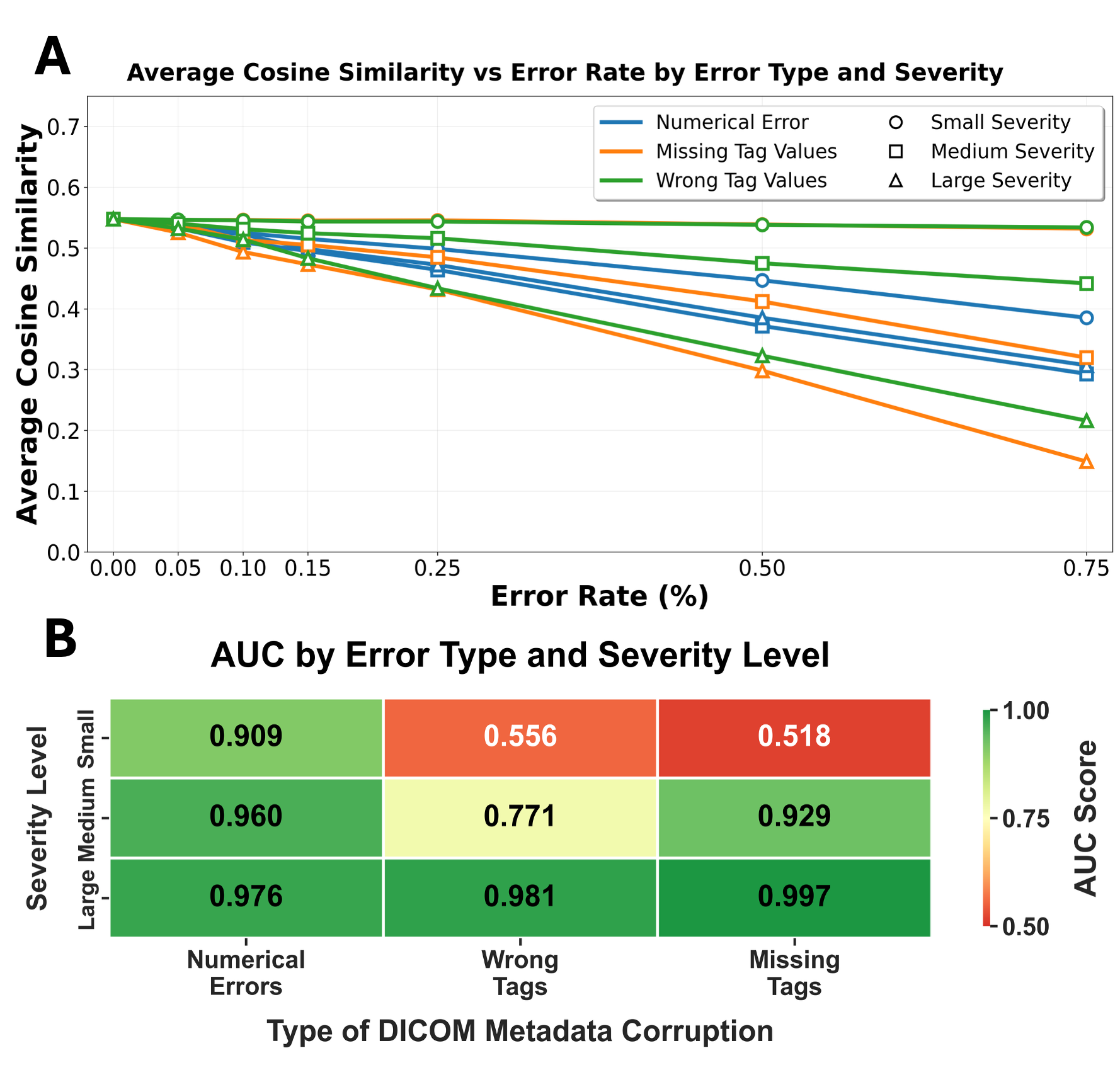}}
  \medskip
\end{minipage}
\caption{Evaluation of quality control, showing the degradation of average cosine similarity with increasing metadata error rate (A) and the AUC performance (B) across three error types and severity levels with 50\% error rate.}
\label{fig:qc}
\end{figure}

For unsupervised QC, MR-CLIP evaluates metadata integrity by comparing image–metadata embedding similarity under controlled corruption, where a defined portion of the test set is systematically corrupted as outlined in Table \ref{tab:qc_methodology}. As shown in Fig. \ref{fig:qc}A, similarity consistently decreases with higher corruption rates, demonstrating the model’s strong sensitivity to metadata inconsistencies. Missing tag values have the most pronounced effect, particularly when sequence and scanner fields are absent. In contrast, corruptions in the \textit{Series Description} tag have minimal impact, since this field is inherently noisy and not used for label construction. Fig. \ref{fig:qc}B summarizes detection performance using AUC scores at a 50\% corruption rate. MR-CLIP achieves near-perfect detection for missing categorical tags (AUC = 0.997) and large numerical errors (AUC = 0.976). In contrast, incorrect categorical tags remain more challenging to detect due to their partial semantic alignment with the image.


\section{Discussion}
\label{sec:discussion}

Our results demonstrate that MR-CLIP effectively learns joint image–metadata representations that capture acquisition semantics. The model achieves high linear-probe accuracy across key DICOM fields, robust clustering in latent space, and strong transferability in few-shot sequence recognition. Importantly, its sensitivity to metadata corruptions confirms MR-CLIP’s potential as a practical tool for automated quality control in large-scale MRI repositories. Future work should explore the performance of MR-CLIP on anatomies other than the brain and on multi-site data.



\section{Compliance with Ethical Standards}
Data usage is approved under HRA Generic Approval 
(IRAS ID:349531 REC Reference: 24/ES/0099).

\section{Acknowledgments}
\label{sec:acknowledgments}

This work was supported by the UK EPSRC [EP/Y035216/1] through the DRIVE-Health CDT at King’s College London, with additional support from deepc GmbH and the Scientific and Technological Research Council of Türkiye (TÜBİTAK) 2213-A Overseas Graduate Scholarship.

\bibliographystyle{IEEEbib}
\bibliography{strings,refs}

\end{document}